\documentclass[10pt,twocolumn,letterpaper]{article}

\usepackage{cvpr}
\usepackage{times}
\usepackage{epsfig}
\usepackage{graphicx}
\usepackage{amsmath}
\usepackage{amssymb}
\usepackage{graphicx,import}

\usepackage[pagebackref=true,breaklinks=true,letterpaper=true,colorlinks,bookmarks=false]{hyperref}

\cvprfinalcopy 

\def\cvprPaperID{****} 

\begin{document}

\title{clcNet: Improving the Efficiency of Convolutional Neural Network using Channel Local Convolutions}

\author{Dong-Qing Zhang\\
ImaginationAI LLC\\
{\tt\small dongqing@gmail.com}
}

\maketitle

\begin{abstract}
Depthwise convolution and grouped convolution has been successfully applied to improve the efficiency of convolutional neural network (CNN). We suggest that these models can be considered as special cases of a generalized convolution operation, named channel local convolution(CLC), where an output channel is computed using a subset of the input channels. This definition entails computation dependency relations between input and output channels, which can be represented by a channel dependency graph(CDG). By modifying the CDG of grouped convolution, a new CLC kernel named interlaced grouped convolution (IGC) is created. Stacking IGC and GC kernels results in a convolution block (named CLC Block) for approximating regular convolution. By resorting to the CDG as an analysis tool, we derive the rule for setting the meta-parameters of IGC and GC and the framework for minimizing the computational cost. A new CNN model named clcNet is then constructed using CLC blocks, which shows significantly higher computational efficiency and fewer parameters compared to state-of-the-art networks, when being tested using the ImageNet-1K dataset. 
\end{abstract}

\section{Introduction}

Convolutional Neural Network has achieved tremendous success for many computer vision problems, such as image classification\cite{alexnet2012krizhevsky}, object detection\cite{rcnn2014girshick}, and image segmentation\cite{fcn2015long}. More recently, due to the pervasive use of mobile and wearable devices, and the rise of emerging applications such as self-driving car, CNN models have been implemented in  resource constrained environments, such as mobile and embedded platforms. The computational and memory efficiency of CNN is crucial for its successful deployment on these platforms, since they generally have very strict resource requirements. Therefore, how to improve the computational and memory efficiency of CNN has become an important research topic in the field of deep learning, and also is the focus of this paper.

Convolution layer is the fundamental building block in a convolutional neural network, and was inspired by the model proposed by Hubel and Wiessel \cite{receptivefield1968hubel}, which shows that visual neurons respond to a local small region of the visual field. This leads to one of the central tenets of the convolution operation in CNN: spatial locality, namely, the computation at a cell only involves a local spatial area in the input.  On the other hand, regular convolution always assume that all channels (i.e. the feature planes) of the input are involved in the computation at a cell. Namely, the computation is always global for the channel dimension. It was only unitl recently that researchers started to use depthwise convolution \cite{xception2017chollet} and grouped convolution \cite{alexnet2012krizhevsky} , where computation only involves subsets of the input channels. 

\begin{figure}[t!]
\begin{center}
\includegraphics[width=1.0\linewidth]
                   {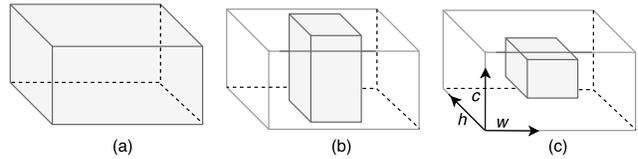}
\end{center}
   \caption{Receptive field (gray-colored cube) in the input tensor of (a) fully connected layer (b) regular convolution (c) channel local convolution }
\label{fig:recept_fd}
\end{figure}

This paper attempts to provide a generalized view about depthwise convolution and grouped convolution through a concept named \textit{channel local convolution}, where the computation of an output channel only depends on a subset of its input channels. Distinct from the regular convolution, the receptive field of channel local convolution is local both along the spatial and channel dimension, which is conceptually illustrated in Figure \ref{fig:recept_fd}. 

A channel local convolution (CLC) kernel is characterized by its \textit{channel dependency graph} (CDG), which is an acyclic graph where the nodes represent channels and edges represent dependencies. CDG describes the computation dependency of the channels, and can be used to analyze a convolution block composed of multiple CLC kernels for approximating regular convolution. 

\textit{Channel receptive field} is another proposed concept analogous to spatial receptive field. When the output channels of convolution depends on all of the input channels, we say that the convolution kernel achieves \textit{full channel receptive field}. By analyzing previous models using grouped convolution and depthwise convolution, we postulate that \textit{full channel receptive field}(FCRF) is necessary for achieving accurate approximation to regular convolution for a  convolution block created by stacking multiple CLC kernels. 

We designed a new convolution block structurally similar to depthwise separable convolution \cite{xception2017chollet} but using two CLC kernels, grouped convolution (GC) and a GC variant named \textit{interlaced grouped convolution(IGC)},  as building blocks. Using the channel dependency graph as the analysis tool, we derive the rule for setting the meta-parameters of the CLC kernels, and present a cost minimization framework for finding the best meta-parameters using the rule as a constraint.

A new convolutional neural network named clcNet is then constructed using the developed CLC blocks. This network is then tested using the ImageNet-1K classification dataset (i.e. the ILSVRC 2012\cite{ilsvrc2015olga} classification dataset). The experiment shows that the clcNet achieves significant computational efficiency improvement and parameter reduction compared to the  state-of-the-art models, while achieving comparable or better accuracy. For example, compared to MobileNet \cite{mobilenet2017howard}, one of the trained clcNet models achieves 25\% reduction in computation with 1.0\% increase of top-1 classification accuracy.
\section{Prior Work}

The research on CNN efficiency improvement can be dated back to early days of convolutional network research. For example, in the work of \textit{optimal brain damage}\cite{prune1990Lecun1990}, the convolution kernel weights are pruned by estimating their contributions to the final loss. The weight pruning approach finds its reincarnation in more recent work such as \cite{compression2015song} with the help of more efficient sparse matrix libraries. However, modern CPU architecture often favors continuous memory addressing and computation. Therefore, weight pruning approaches may be only effective when the weight matrix is sufficiently sparse.

For this reason, more recent work has been focused on uniform sparsification of convolution kernel or filter-level sparsification. A typical example is the \textit{Inception} module in GoogleNet \cite{inception2015szegedy}, which seeks optimal sparse structure of convolution by searching the best mixture of small and large convolution kernels. Similar idea is also adopted in SqueezeNet\cite{squeezenet2016iandola}, where the \textit{fire module }is constructed by mixing 3x3 and 1x1 kernels. Another popular CNN model is Residual Newtwork \cite{resnet2016he}, in which the residual block can be considered as the sparsification of regular convolution by the summation of an identity function and a residual function. The $L_2$ weight regularization (weight decay) then would make the residual function much closer to zero. The ResNeXt model\cite{resnext2016saining} further extends the residual transform idea to the use of multi-branch residual functions, which can be implemented using grouped convolution. From the perspective of the weight matrix of a convolution kernel, the grouped convolution is manifested as block diagonal matrix. 

Another large category of approaches is low-rank approxiamtion \cite{lowrank2014denton}\cite{lowrank2014jaderberg}\cite{lowrank2014labedev}, where convolution tensor is approximated by decomposing it into the composition of smaller low-rank tensors. This decompsotion can be achieved in different manners, for instance, by decompsing 4D tensor into product of two 3D tensors \cite{lowrank2014jaderberg}, by SVD or outer-product decomposition \cite{lowrank2014denton}, or by using direct CP-decomposition \cite{lowrank2014labedev}. 

Other indirectly related work include Fast Fourier Transform \cite{fft2017mathieu}, convolution weight quantization\cite{compression2015song}, and binary weight network\cite{bin_net2016rastegari}.

The proposed method is most related  to two recent work: depthwise separable convolution \cite{xception2017chollet}\cite{mobilenet2017howard}\cite{depth_separable2016wang} and ShuffleNet\cite{shufflenet2017xiangyu}. In  \cite{xception2017chollet}\cite{mobilenet2017howard}, depthwise separable convolution is created by stacking 3x3 depthwise convolution with 1x1 pointwise convolution. This achieves large computation and parameter reduction. However, the 1x1 pointwise convolution dominates the computational cost and parameter count in depth seperable convolution, making further efficiency gain difficult unless we partition the 1x1 pointwise convolution further. 

The work of ShuffleNet\cite{shufflenet2017xiangyu} attempts to overcome this limitation by using grouped convolution with channel shuffling. Its is based on convolution blocks following a three-layer sandwich-like structure more similar to those in the ResNeXt\cite{resnext2016saining} model. However, although it achieves significant efficiency improvement for AlexNet\cite{alexnet2012krizhevsky} level classification accuracy, its efficiency improvement is less favorable for higher classification accuracy(e.g. above 70\% top-1 accuracy). The \textit{interlaced grouped convolution}, which is part of the proposed CLC convolution block in this paper, may also be implemented as group convolution followed by channel shuffling. However, the proposed CLC block is a simpler two-layer structure. This simplified design requires more careful choices of the convolution meta-parameters in order to achieve minimal cost and accurate approximation to regular convolution. And the proposed framework provides a systematic way and optimization-based framework for choosing the best meta-parameters, leading to lower computational cost of the proposed clcNet compared to ShuffleNet. In comparison, the ShuffleNet work does not provide similar framework for general designs, such as the proposed two-layer structure.


\section{Channel Local Convolution}

The idea of channel local convoltuion is inspired by grouped convolution. In grouped convolution, the computation at a location only needs the input channels that belong to the same group as the output channel. Grouped convolution was first adopted in \cite{alexnet2012krizhevsky} for distributing convolution computation to multiple GPUs.  Later, it was utilized by ResNeXt\cite{resnext2016saining} model for implementing multi-branch residual transform. The success of ResNeXt network implies that the convolution operation can be sparse along the channel dimension while still achieving high representation power. Such channel-wise sparsity reduces computational cost and number of parameters. And fewer parameter leads to higher generalization capability for classification. 

Similarly, the depthwise convolution in depth separable convolution is a grouped convolution with one group. It leads to drastically reduced parameter count when being combined with pointwise convolution in the MobileNet\cite{mobilenet2017howard} and Xception\cite{xception2017chollet} model.

The grouped convolution has a distinct channel dependency pattern: an output channel is only dependent on the input channels in the same group.But can we have convolution kernels with channel dependency not being confined in the groups? This question inspires us to extend grouped convolution to a more general concept, named \textit{channel local convolution} (CLC), where an output channel can depend on an arbitrary subset of the input channels. 

Formally, we define \textit{channel local convolution }as a convolution operation where an output channel is computed using a subset of the input channels. This definition does not exclude the possibility that the dependent input channels of an output channel are scattered along the channel dimension. However, we expect that it is most often that the dependent input channels are neighboring along the channel dimension (e.g. in grouped convolution), forming local regions in the channel domain. 

\subsection{Channel Dependency Graph and Channel Receptive Field}

\begin{figure}[t!]
\begin{center}
\includegraphics[width=1.0\linewidth]
                   {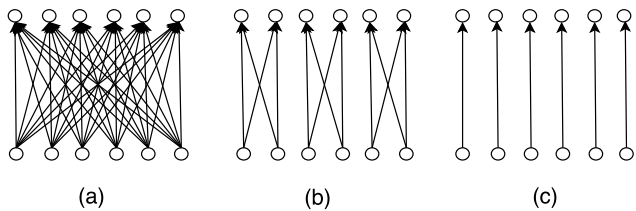}
\end{center}
\caption{Channel dependency graph of (a) regular convolution, (b) grouped convolution, (c) depthwise convolution. Note: The arrows of edges point from output channel nodes to their dependent input channel nodes}
\label{fig:clc}
\end{figure}

A channel local convolution(CLC) kernel is characterized by its \textit{channel dependency graph}(CDG). CDG is a directed acyclic graph, where nodes represent channels, and edges represent channel dependency relations. For a CLC kernel, its CDG is a bipartite graph, where the top row of nodes represent the input channels, and the bottom row represents the output channels. The arrow of an edge points from an output channel node to its corresponding input channel node. Figure \ref{fig:clc} illustrates the CDGs of regular convolution, grouped convolution and depthwise convolution. CDG can be used to facilitate the analysis of channel dependency for designing convolution blocks composed of multiple CLC kernels.

Similar to the concept of spatial receptive field, we define the \textit{channel receptive field } (CRF) of an output channel in a convolution kernel or block as the subset of the input channels that the output channel depends on. 
Similar to the concept of receptive field size in spatial domain, the \textit{channel receptive field size} (CRF size) of an output channel is defined as the number of the dependent input channels of the output channel. If every output channel in a convolution kernel or block has the same CRF size $s$, then we say the convolution kernel or block has CRF size of $s$. It can be observed that grouped convolution has CRF size of $M/g$, where $M$ is the number of input channels, and $g$ the group parameter (number of groups). And depthwise convolution has CRF size of 1. For regular convolution, its CRF size is $M$. A regular convolution kernel can be approximated by a convolution block composed of multiple convolution kernels. For instance, depthwise separable convolution is a convolution block by stacking 3x3 depthwise convolution and 1x1 pointwise convolution. If a convolution kernel or block has its CRF size equals to the number of its input channels, we say the convolution kernel or block has \textit{full channel receptive field} (FCRF). 

\begin{figure}[b!]
\begin{center}
\includegraphics[width=1.0\linewidth]
                   {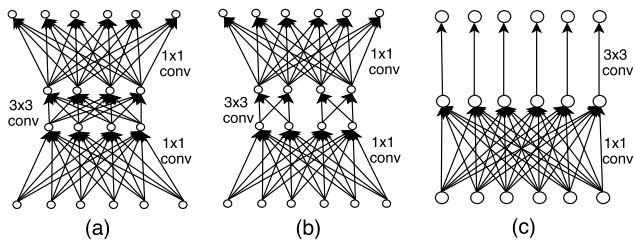}
\end{center}
   \caption{Channel dependency graph of convolution blocks (a) ResNet bottleneck structure\cite{resnet2016he}, (b) ResNeXt block\cite{resnext2016saining}, (c) Depth separable convolution in MobileNet\cite{mobilenet2017howard} and Xception\cite{xception2017chollet} }
\label{fig:fcrf}
\end{figure}

We postulate that in order for a convolution block to achieve accurate approximation to regular convolution, it needs to attain full channel receptive field(FCRF). We argue that this is because the regular convolution has FCRF, and not achieving FCRF would result in fewer effective channels for  feature representation, leading to smaller effective network width. And prior work\cite{wideresnet2016Zagoruyko} has demonstrated that larger network width improves representation power similar to increasing depth. Our experiments also validate that FCRF is critical to achieve high classification accuracy. Large accuracy degradation would be observed if FCRF is not achieved for the convolution blocks in the network.

Channel dependency graph (CDG) can be used to analyze a convolution block to verify if it achieves FCRF and facilitate the design of a convolution block for achieving FCRF.  Figure \ref{fig:fcrf} illustrates the CDGs of the convolution blocks propsed in the previous work, including the bottleneck structure in ResNet \cite{resnet2016he},  the ResNext block \cite{resnext2016saining}, and depth-separable convolution \cite{mobilenet2017howard}. It can be observed that all of them achieve FCRF.

\subsection{Interlaced Grouped Convolution (IGC) and CLC Block}

The depth separable convolution, used by MobileNet\cite{mobilenet2017howard} and Xception\cite{xception2017chollet}, has been proved very efficient for convolutional neural network. However, the computational cost of depth separable convolution is dominated by the pointwise convolution. Further cost reduction can be only achieved by partitioning the pointwise convolution, for example using grouped convolution.

\begin{figure}[t!]
\begin{center}
\includegraphics[width=0.85\linewidth]
                   {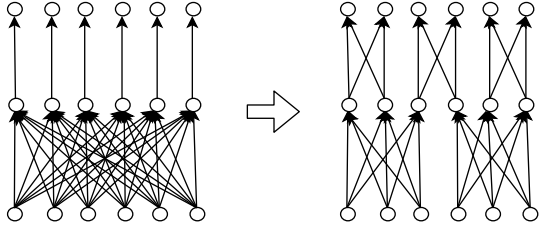}
\end{center}
   \caption{Replacing pointwise and depthwise convolution with grouped convolution results in loss of full channel receptive field property}
\label{fig:lost_FCRF}
\end{figure}

Nevertheless, our initial experiment shows that naively replacing pointwise and depthwise convolution with grouped convolution results in large degradation of classification accuracy. If we look at the CDG of the modified block, it is evident that the full channel receptive field (FCRF) property is lost (Figure \ref{fig:lost_FCRF}).

\begin{figure}[b!]
\begin{center}
\includegraphics[width=0.6\linewidth]
                   {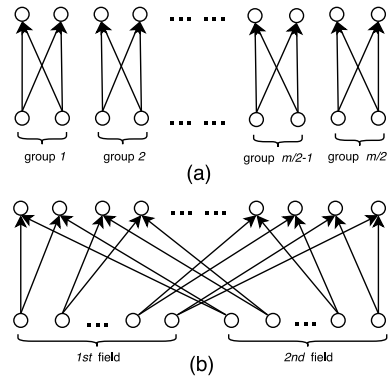}
                   
\end{center}
   \caption{Example channel dependency graph of (a)regular grouped convolution, (b) interlaced grouped convolution}
\label{fig:igc}
\end{figure}

To remedy the above problem, we can change the channel dependency pattern of one of the grouped convolutions by keeping the channel connectivity in its CDG unchanged but interlacing the output channels into a number of "fields". This results in a special case of channel local convolution with the altered CDG shown in Figure \ref{fig:igc}(b). The CDG creation process is analogous to that of the interlaced video format used in broadcast TV industry, therefore the convolution operation is named \textit{interlaced grouped convolution} (IGC). Similar to grouped convolution, IGC computation can be performed group by group. Therefore, IGC can also be parameterized by the group parameter $g$. And the channel receptive field size of IGC is $M/g$, where $M$ is the number of input channels. The number of the partitioned "fields" is equal to the channel receptive field size. 

A \textit{CLC block} is a convolution block constructed by stacking a 3x3 interlaced grouped convolution(IGC) kenrel with a 1x1 grouped convolution(GC) kernel, with additinal ReLU activiation and batch normalization layers. Similar to the Xception model \cite{xception2017chollet}, there is no activation function following the IGC kernel. The structure is illustrated in Figure \ref{fig:FCRF} (left).

\subsection{Rule for FCRF and Cost Minimization}
The CLC block can achieve full channel receptive field (FCRF) if the group parameter $g$ of the  IGC and GC  kernel is set properly. 

Figure \ref{fig:FCRF} illustrates the CLC block (left) and its channel dependency graph (right). In the CDG, the block has $M$ input channels, $N$ output channels, and $L$ intermediate channels (the output from the IGC kernel). The group parameter of the IGC kernel and GC kernel is $g_1$ and $g_2$ respectively. It can be seen that the IGC kernel has channel receptive filed size (CRF size) $M/g_1$, and the GC kernel has CRF size $L/g_2$. In order for the output channel having the receptive field covering all the input channels of the block, we need to have $L/g_2\ge{g_1}$ or $g_1g_2\le{L}$. Therefore, we can summarize the condition to set the meta-parameters $g_1$ and $g_2$ for achieving FCRF as the following :

\begin{figure}[t!]
\begin{center}
\includegraphics[width=1.0\linewidth]
                   {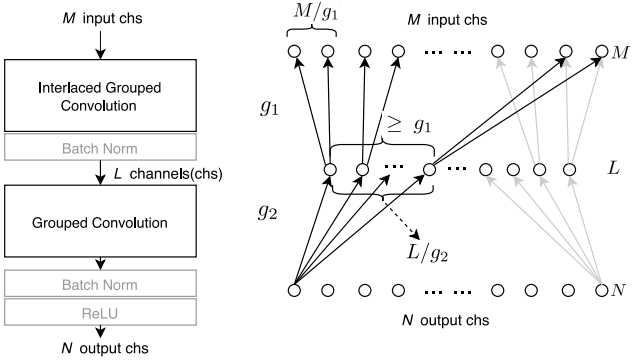}
\end{center}
   \caption{CLC block and its channel dependency graph}
\label{fig:FCRF}
\end{figure}

\vspace{10pt}
\textbf{\textit{Rule for FCRF}}: \textit{if a convolution block is constructed by stacking an IGC kernel with a GC kernel, to achieve full channel receptive field (FCRF) property for the block, the group parameter $g_1$ of the IGC kernel and $g_2$ of the GC kernel have to satisfy the following condition:} 
\begin{equation}
g_1g_2 \le{L}
\label{eq:FCRF}
\end{equation}
\textit{Where $L$ is the number of output channel of the IGC kernel.}
\vspace{4pt}

Based on \textit{the rule for FCRF}, every CLC block should have a lower bound of computational cost to achieve FCRF.  And the group parameters to achieve the lower bound can be found by minimizing the computational cost per location (equals to number of parameters) with \textit{the rule for FCRF} as inequality constraint, which is written as the following:

\begin{align}
& \underset{g_1,g_2}{\text{minimize}} & & C(g_1,g_2) = \frac{ALM}{g_1} + \frac{NL}{g_2} \label{eq:min_cost} \\
\notag& \text{subject to} & & 1\le{g_1}\le{\min(M,L)},\  1\le{g_2}\le{\min(L,N)}, \\
\notag&&& M,L \bmod g_1=0,\ \ L,N\bmod g_2=0, \\
\notag&&& g_1g_2\le{L}.
\end{align}

Where $A$ is the spatial area of the convolution kernel. For instance, $A$ equals 9 for a 3x3 kernel. Note that the above equation assumes that the IGC kernel is 3x3 and GC kernel is 1x1. Similar equations can be derived for other cases, for instance for 3x3 GC kernel. It also assumes that \textit{stride} equals 1 for the GC kernel. For the above minimization problem,  since the group parameters are discrete and lie in a limited range, it can be simply solved by enumerating all possible values of $g1$ and $g2$. Table \ref{tb:fcrf} shows a list of minimization results for typical input and output channels when $A=9$.
\begin{table}[h!]\scriptsize
\begin{center}
\begin{tabular}{|l|l|l||c|c|||l|l|l||c|c|}
\hline
$M$ & $L$ & $N$ & $g_1$ & $g_2$ & $M$ & $L$ & $N$ & $g_1$ & $g_2$ \\
\hline\hline
32 & 32 & 64 & 16 & 2 & 64 & 64 & 64 & 32 & 2 \\
\hline
64 & 64 & 128 & 16 & 4 & 128 & 128 & 128 & 32 & 4 \\
\hline
128 & 128 & 256 & 32 & 4 & 256 & 256 & 256 & 64 & 4 \\
\hline
256 & 256 & 512 & 32 & 8 & 512 & 512 & 512 & 64 & 8 \\
\hline
512 & 512 & 1024 & 64 & 8 & 1024 & 1024 & 1024 & 128 & 8 \\
\hline
\end{tabular}
\end{center}
\caption{Minimization results for typical input and output channels}
\label{tb:fcrf}
\end{table}

\section{The clcNet}
A new convolutional network, named  \textit{clcNet}, is constructed using the CLC blocks. The macro structure of this network is roughly the same as MobileNet, but all the depth separable convolution layers are replaced by CLC blocks, and the number of CLC blocks in different stages of the network is changed for getting various classification accuracy. 

\subsection{Network Design}

The group parameters $g_1$ and $g_2$ have to be determined for every CLC block when designing clcNet. In theory, we can set $g_1$ and $g_2$  to be the values that achieve the cost lower bound. However, allocating too few parameters to the late layers of the network would result in large degradation of classification accuracy, while contributing not much to the overall cost reduction, because the computational cost is usually concentrated at early layers. 

Another consideration is about the implementation of the IGC kernel. If the channel receptive field size is small, the IGC kernel could be more efficiently implemented using depthwise convolution, because a specialized implementation (e.g. in TensorFlow) of depthwise convolution could be faster than regular grouped convolution implementation. Due to these considerations, we fix the $g_2$ parameter to 2 for all CLC blocks, and find the value of $g_1$ using Eq.(\ref{eq:min_cost}) to minimize the computational cost. This can achieve computational cost close to the lower bound for early layers, and make the channel receptive field size of the IGC kernel equal to 2.

The overall design of the network is shown in Table \ref{tb:clcnet}. For all the CLC blocks, we set $L=M$, therefore the group parameters are only determined by the number of input and output channels of the blocks. The parameter $a$,$b$,$c$,$d$ in the table is the count of CLC block repetition at different stages in the network. Changing them would vary the accuracy and computational cost of the network.

\begin{table}[h!]\footnotesize
\begin{center}
\begin{tabular}{|c||c|c|c||c|c|}
\hline
Block type & Input & Output  & Stride & $g_1$  & $g_2$  \\
\& repetition & channel & channel & &   &   \\
\hline\hline
Regular 3x3 conv & 3 & 32 & 2 &   &   \\
\hline
\multicolumn{6}{|c|}{BatchNorm+ReLU}\\
\hline
CLC block & 32 & 64 & 1 & 16  & 2  \\
\hline\hline
CLC block & 64 & 128 & 2 & 32  & 2  \\
\hline
CLC block$\times{a}$ & 128 & 128 & 1 & 64  & 2  \\
\hline\hline
CLC block & 128 & 256 & 2 & 64  & 2  \\
\hline
CLC block$\times{b}$ & 256 & 256 & 1 & 128  & 2  \\
\hline\hline
CLC block & 256 & 512 & 2 & 128  & 2  \\
\hline
CLC block$\times{c}$ & 512 & 512 & 1 & 256  & 2  \\
\hline\hline
CLC block & 512 & 1024 & 2 & 256  & 2  \\
\hline
CLC block$\times{d}$& 1024 & 1024 & 1 & 512  & 2  \\
\hline
\multicolumn{6}{|c|}{Average Pooling}\\
\hline
FC layer & 1024  & 1000  &  &   &   \\
\hline
\end{tabular}
\end{center}
\caption{The structure and block parameters of the clcNet, where $a,b,c,d$ is the count of block repetition, which can be changed for different performance}
\label{tb:clcnet}
\end{table}

\subsection{Network Implementation Issues}

For interlaced grouped convolution (IGC), although we can have a two-step implementation composed of grouped convolution and channel interlacing, the preferable way for production deployment is to have a monolithic implementation that directly access the respective channels. 

For prototyping or experiment purposes, the IGC may be more easily implemented using built-in components with custom layers. For example, on Torch\cite{torch2011collobert}, PyTorch or Caffe\cite{caffe2014jia} platform, there is a built-in implementation of grouped convolution. Thus, IGC can be implemented with grouped convolution followed by a custom channel interlacing layer. This is the choice of our current implementation.

On the Tensorflow\cite{tensorflow2016abadi} platform, there is no existing grouped convolution component. But if the channel receptive field of IGC is small, it can be implemented using depthwise convolution. More specifically, if the channel receptive field size is 2, the IGC kernel can be implemented with two depthwise convolution operations, one acting on the original input, and the other on the input with its odd and even channels switched.

\section{Experiments}

The experiments are intended to evaluate the effectiveness of the clcNet, and its computational efficiency compared to state-of-the-art models with comparable image classification accuracy. 

The clcNet is implemented on Torch platform using the codebase \cite{fb.resnet} for ResNet implementation provided by Facebook AI Research (FAIR). The IGC kernel is implemented using grouped convolution followed by a custom channel interlacing layer.

The experiments are conducted on ImageNet-1K dataset (a.k.a ILSVRC 2012 image classification dataset) to evaluate the top-1 and top-5 image classification accuracy. Details of this dataset can be found in \cite{ilsvrc2015olga}. The same as prior work, the validation dataset is used as a proxy to test set for accuracy evaluation. Previous papers, for instance \cite{resnet2016he}, have shown that the cross-experiment variation of test accuracy is very small for the ImageNet-1K dataset due to its large size, compared with other smaller-sized datasets. Therefore, only ImageNet-1K dataset is used for evaluation.

The learning optimizer is important to ensure best accuracy results.  SGD and RMSProp\cite{rmsprop2012tieleman} are two popular optimizers used by previous work. In our experiments, both SGD and RMSprop optimizers are tested. And SGD is found to result in a slightly better accuracy than RMSProp. The SGD optimizer uses the default setting in the ResNet codebase, where momentum is set to 0.9, and \textit{nesterov} momentum is enabled. As to learning rate schedule, a polynomial learning rate schedule is used, where the \textit{power} parameter is set to 1.0, which means a linear decay of learning rate. The polynomial schedule is used, because our initial experiment shows that it can reproduce MobileNet's accuracy reported in \cite{mobilenet2017howard}, while the default multi-step schedule (times 0.1 per 30 epochs) in ResNet codebase cannot. The initial learning rate is set to 0.1. And the learning process runs for 100 epochs. 

The regularization parameter, namely \textit{weight decay}, is another important factor for achieving the optimal result. We did not run an extensive search for finding optimal weight decay due to lack of resource. Only two weight decay values are experimented: $4.0\mathrm{e}{-5}$ and $1.0\mathrm{e}{-4}$, where the former is used by the Inception model\cite{inception2015szegedy}, and latter by the ResNet model\cite{resnet2016he}.  The value of $1.0\mathrm{e}{-4}$ is found to have a better result. Also, unlike in MobileNet where weight decay is set to different values for depthwise and pointwise convolution, the weight decay is the same for all convolution kernels in training clcNet.

For data augmentation, we use the default data augmentation module in the ResNet codebase, which performs randomized crop, color jittering, and horizontal flips to the training images. All the training and testing images are resized and cropped to the size of $224\times{224}$. And for testing time, only single crop and single model evaluation is performed. The image preprocessing process for evaluation uses the default settings in the ResNet codebase. 

\subsection{Classification Accuracy of clcNet}

For comparing with state-of-the-art models, we tried different layer configurations ($a$,$b$,$c$,$d$ value in Table \ref{tb:clcnet}) of the network for matching the top-1 classification accuracy with MobileNet. At the end, two clcNets with different configurations are chosen. They are named clcNet-A and clcNet-B respectively. The configurations and classification accuracy of these two networks are shown in Table \ref{tb:exp_try} below. And Figure \ref{fig:train_profile} shows the evolution of the top-1 test accuracy on the validation dataset during the training process for clcNet-A and clcNet-B.
\begin{table}[h!]
\begin{center}
\begin{tabular}{l|c|c|c}
\hline
Model & (a,b,c,d) & Top-1 Acc.  & Top-5  Acc.\\
\hline\hline
clcNet-A  & $(1,1,5,2)$ & 70.4\%  & 89.5\% \\
clcNet-B  & $(1,1,7,3)$ & 71.6\%  & 90.3\% \\
\hline
\end{tabular}
\end{center}
\caption{Classification accuracy of clcNet on ImageNet-1K }
\label{tb:exp_try}
\end{table}
\begin{figure}[h!]
\begin{center}
\includegraphics[width=1.0\linewidth]
                   {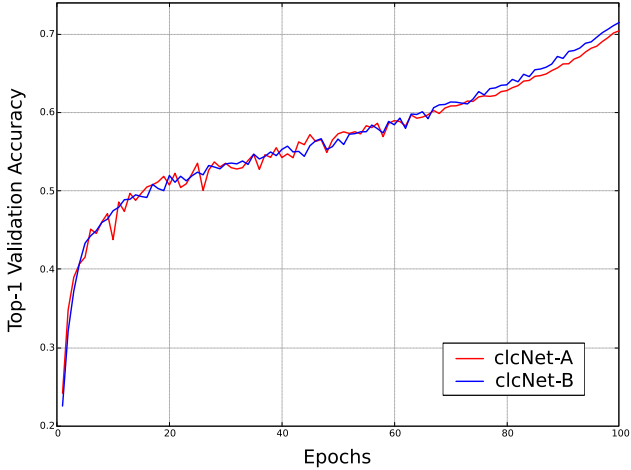}
\end{center}
   \caption{Training profile on ImageNet-1K dataset }
\label{fig:train_profile}
\end{figure}

Our experiments with different layer configurations suggest that adding more late layers (i.e. increasing $c$ or $d$) results in accuracy increase faster than adding more early layers ($a$ or $b$), with more parameter increase but less computational efficiency degradation.

\subsection{The importance of FCRF}
To verify the importance of the full channel receptive field(FCRF) property, we replace all the IGC kernels in clcNet-A to GC kernels, resulting in a network (clcNet-Ap) with almost the same computational cost but loss of FCRF for CLC blocks. Table \ref{tb:fcrf} shows the result comparison of clcNet-A and its modified version (clcNet-Ap). The results demonstrate the importance of the FCRF property.
\begin{table}[h!]
\begin{center}
\begin{tabular}{l|c|c}
\hline
Model &  Top-1 Acc.  & Top-5  Acc.\\
\hline\hline
clcNet-A  &  70.4\%  & 89.5\% \\
clcNet-Ap &  67.7\%  & 87.8\% \\
\hline
\end{tabular}
\end{center}
\caption{Comparison of the results of clcNet-A and clcNet-Ap}
\label{tb:fcrf}
\end{table}
\subsection{Comparison with Previous Models}
Because our model is targeted for resource constraint environments, such as mobile platform, we only compare with the previous models with low computational cost, and small memory footprint, which entails small model size and network width. Both MobileNet\cite{mobilenet2017howard} and ShuffleNet\cite{shufflenet2017xiangyu} are designed for mobile platforms, therefore they are selected for comparison baselines.

The metrics for comparison include top-1 accuracy, mult-add operation count and number of parameters. These are also used in previous papers for benchmark.  Table \ref{tb:exp_comp} lists the performance comparison between the previous models and the clcNets, including clcNet-A and clcNet-B.

\begin{table}[h!]\small
\begin{center}
\begin{tabular}{l|c|c|c}
\hline
Model  & Top-1 Acc. & Mult-Adds & Parameters \\
\hline\hline
GoogleNet  & 69.8\%  & 1550M & 6.8M \\
1.0 MobileNet-224  & 70.6\%  & 569M & 4.2M \\
ShuffleNet 2$\times$  & 70.9\% (v1)  & 524M & 5.3M \\
\hline
\textbf{clcNet-A}  & \textbf{70.4\%}  & \textbf{343M} & \textbf{3.25M} \\
\textbf{clcNet-B}  & \textbf{71.6\%}  & \textbf{425M} & \textbf{4.1M} \\
\hline
\end{tabular}
\end{center}
\caption{Comparison with previous models for classification accuracy and computational cost}
\label{tb:exp_comp}
\end{table}

It can be observed that the developed clcNet models achieve significant improvements for reducing computational cost and parameter count compared to previous models. More specifically, 
compared to MobileNet, the clcNet-A model achieves 40\% reduction in computation and 22.6\% reduction in parameter count with a slightly lower top-1 accuracy. And the clcNet-B model achieves 25\% reduction in computation with 1.0\% increase of top-1 accuracy. Compared to ShuffleNet (v1\cite{shufflenet2017xiangyu}), the clcNet-B mdoel achieves 19\% reduction in computation, and 18\% fewer parameters with 0.7\% increase of top-1 accuracy.  


The computational cost in Table \ref{tb:exp_comp} is measured by multiply-add operations(MACs), which does not consider other costs such as memory accessing. The cache miss during memory read could become a significant overhead for overall computation. Therefore, it is also important to compare the model inference speed running on actual devices. However, the actual inference speed highly depends on the way of implementation of the convolution and other operations. Therefore, it should not be considered alone for cost measurement either. 

We test the actual inference speed of different models on an Android-based smartphone, BLU Advance 5.2, which is a low-cost smartphone model. This device uses a MediaTek MT6580 SOC with a Quad-core, 1300MHz ARM Cortex-A7 processor, and has 1GB RAM. To run the model inference on the smartphone, the PyTorch implementations of the models are converted to the ONNX models, which are further converted to the Caffe2 models that can be run on smartphones using Caffe2 platform. Since this is a general model conversion without specific code optimization, there could be large space to improve the absolute inference speed. But the speed of clcNet relative to other models should be still meaningful for evaluating its advantage. Table \ref{tb:actual_inf} lists the actual inference speed (average of ten inference passes) of different models on the BLU phone, along with their converted ONNX file sizes. It can be observed that clcNets are still able to achieve significant speedup compared to MobileNet and ShuffleNet, although the percentage of speedup is lower than that of the theoretical computational cost listed in Table \ref{tb:exp_comp}. This could be caused by lower cache effiency after partitioning further the 1x1 convolution weight matrix in the clcNet. 

\begin{table}[h!]\small
\begin{center}
\begin{tabular}{l|c|c|c}
\hline
Model  & Mult-Adds & ONNX size & Actual Speed \\
\hline\hline
1.0 MobileNet-224  & 569M & 17.1M & 1780ms\\
ShuffleNet 2$\times$  & 524M & 22.3M  & 1812ms\\
\hline
\textbf{clcNet-A}  & \textbf{343M} & \textbf{17.3M} & \textbf{1431ms}\\
\textbf{clcNet-B}  & \textbf{425M} & \textbf{21.6M} & \textbf{1692ms}\\
\hline
\end{tabular}
\end{center}
\caption{Comparison of the actual inference speed on smartphone}
\label{tb:actual_inf}
\end{table}

\section{Conclusion}
We propose that depthwise convolution and grouped convolution can be viewed as special cases of channel local convolution(CLC). New analysis tools and concepts, such as channel dependency graph and channel receptive field, are introduced to help the analysis and design of CLC models. We then construct a novel convolution block named CLC block, which is composed of two CLC kernels: grouped convolution and interlaced grouped convolution. A new convolutional neural network named clcNet then is constructed using the CLC blocks. The experiments on ImageNet-1K data show that the clcNet achieves significant efficiency improvements on top of state-of-the-art networks. In addition to the contribution of the clcNet, the framework of channel local convolution along with the proposed analysis tools provides a new paradigm for designing more efficient convolution kernels in the future. 

\small
\bibliographystyle{ieee}
\bibliography{egpaper_final.bib}

\end{document}